\documentclass{article}

\usepackage[utf8]{inputenc}
\usepackage{amssymb}
\usepackage{amsmath}
\usepackage[algo2e]{algorithm2e}
\usepackage{booktabs}
\usepackage{hyperref}
\usepackage{algorithm}
\usepackage{graphicx}
\usepackage{subcaption}
\usepackage{enumerate}
\usepackage{tikz}
\usepackage{gensymb}
\usetikzlibrary{arrows}

\def\E{\mathbb{E}}

\usepackage[accepted]{icml2020}

\begin{document}

\twocolumn[

\icmltitle{NGBoost: Natural Gradient Boosting for Probabilistic Prediction}

\icmlsetsymbol{equal}{*}

\begin{icmlauthorlist}
\icmlauthor{Tony Duan}{equal,st}
\icmlauthor{Anand Avati}{equal,st}
\icmlauthor{Daisy Yi Ding}{st}
\icmlauthor{Khanh K. Thai}{kp}
\icmlauthor{Sanjay Basu}{hms}
\icmlauthor{Andrew Ng}{st}
\icmlauthor{Alejandro Schuler}{kp}
\end{icmlauthorlist}

\icmlaffiliation{st}{Stanford University, Stanford, California, United States}
\icmlaffiliation{hms}{Harvard Medical School, Cambridge, Massachusetts, United States}
\icmlaffiliation{kp}{Unlearn.ai, San Francisco, California, United States}

\icmlcorrespondingauthor{Anand Avati}{avati@cs.stanford.edu}
\icmlcorrespondingauthor{Tony Duan}{tonyduan@cs.stanford.edu}
\icmlcorrespondingauthor{Alejandro Schuler}{alejandro.schuler@gmail.com}

\icmlkeywords{gradient boosting, probabilistic regression}

\vskip 0.3in
]

\printAffiliationsAndNotice{\icmlEqualContribution} 

\begin{abstract}
We present Natural Gradient Boosting (NGBoost), an algorithm for generic probabilistic prediction via gradient boosting. Typical regression models return a point estimate, conditional on covariates, but probabilistic regression models output a full probability distribution over the outcome space, conditional on the covariates. This allows for predictive uncertainty estimation ---  crucial in applications like healthcare and weather forecasting. 
NGBoost generalizes gradient boosting to probabilistic regression by treating the parameters of the conditional distribution as targets for a multiparameter boosting algorithm. Furthermore, we show how the \emph{Natural Gradient} is required to correct the training dynamics of our multiparameter boosting approach. NGBoost can be used with any base learner, any family of distributions with continuous parameters, and any scoring rule. NGBoost matches or exceeds the performance of existing methods for probabilistic prediction while offering additional benefits in flexibility, scalability, and usability. An open-source implementation is available at \href{https://github.com/stanfordmlgroup/ngboost}{github.com/stanfordmlgroup/ngboost}.

\end{abstract}

\section{Introduction}

\begin{figure}[t]
    \centering
    \includegraphics[width=\linewidth]{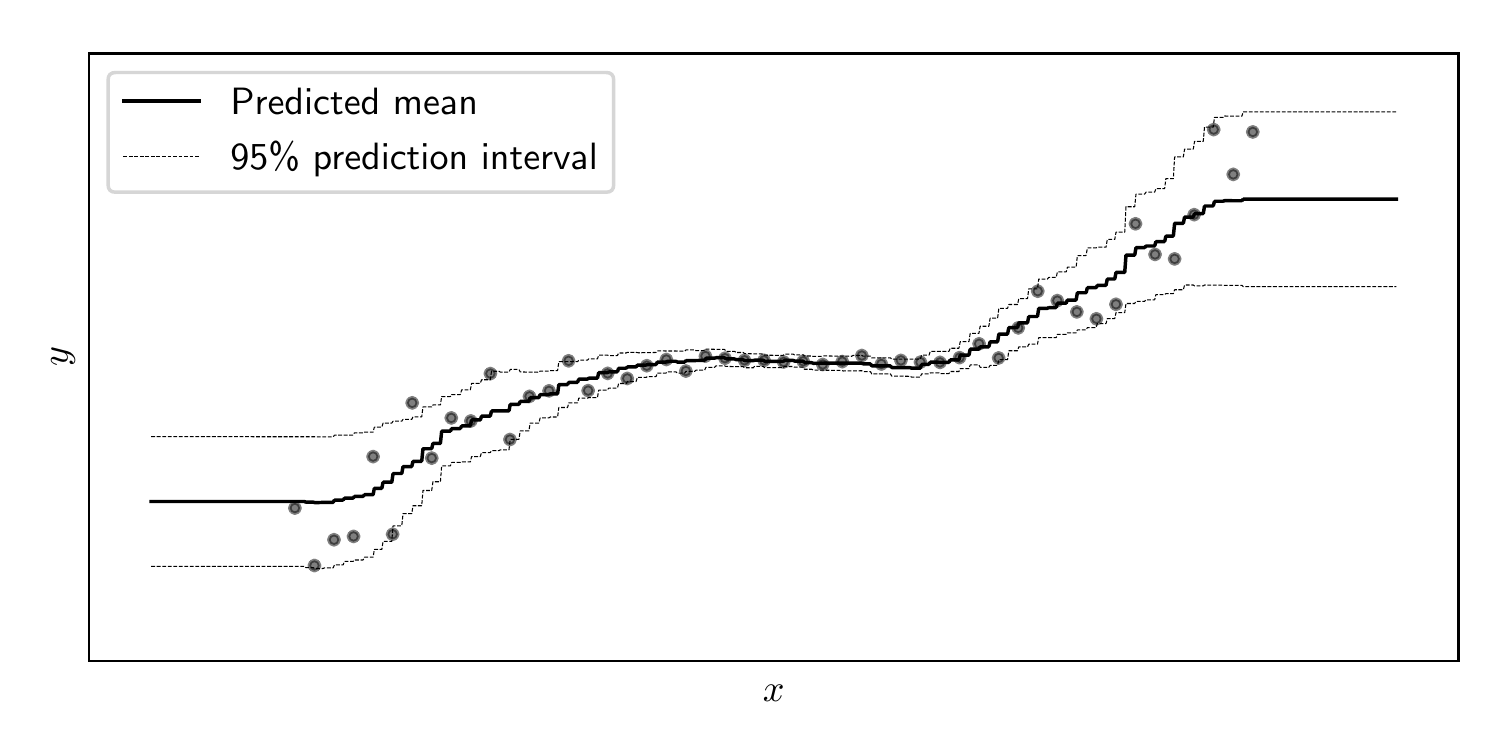}
    \caption{Prediction intervals for a toy 1-dimensional probabilistic regression problem, fit via NGBoost. The dots represent data points. The thick black line is the predicted mean after fitting the model. The thin gray lines are the upper and lower quantiles covering 95\% of the prediction distribution.}
    \label{fig:toy}
\end{figure}

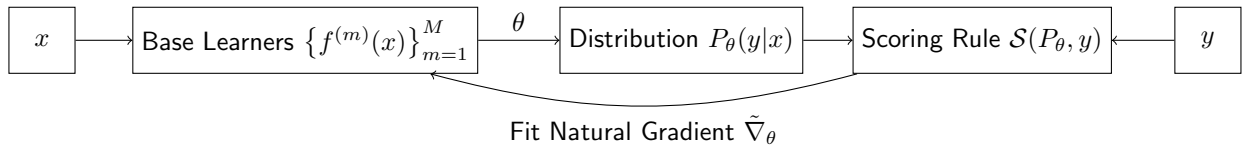
\begin{figure*}[t]
\centering
\begin{tikzpicture}
\tikzset{vertex/.style = {shape=rectangle,draw,minimum size=2.5em}}
\node[vertex] (A) at  (0,0) {$x$};
\node[vertex] (X) at  (3.5,0) {\fontfamily{cmss}\selectfont Base Learners $\left\{f^{(m)}(x)\right\}_{m=1}^M$};
\node[vertex] (Z) at  (8.5,0) {\fontfamily{cmss}\selectfont Distribution $P_\theta(y|x)$};
\node[vertex] (S) at  (12.5,0) {\fontfamily{cmss}\selectfont Scoring Rule $\mathcal{S}(P_\theta, y)$};
\node[vertex] (Y) at  (15.5,0) {$y$};
\draw[->] (A) to (X);
\draw[->] (X) to node[above] {$\theta$} (Z);
\draw[->] (Z) to (S);
\draw[<-] (S) to (Y);
\draw[<-] (X) to [bend right=15] node[below] {\fontfamily{cmss}\selectfont Fit Natural Gradient $\tilde\nabla_\theta$} (S);
\end{tikzpicture}
\caption{NGBoost is modular with respect to choice of base learner, distribution, and scoring rule.}
\label{fig:block_diagram}
\end{figure*}

Many important supervised machine learning problems are regression problems. Weather forecasting (predicting temperature of the next day based on today's atmospheric variables \citep{gneiting_probabilistic_2014}) and clinical prediction (predicting time to mortality with survival prediction on structured medical records of the patient \citep{avati_improving_2018}) are important examples. 

Most machine learning methods tackle this problem with point prediction, returning a single ``best guess'' prediction (e.g. the temperature tomorrow will be 16\degree C). However, in these fields it is often important to be able to quantify uncertainty in the prediction or be able to answer multiple questions on the fly (e.g. what's the probability it will be between 18\degree C and 20\degree C? What about $<$15\degree C?) \citep{ml_meet_econ}. 

In order to answer arbitrary questions about the probability of events conditional on covariates, we must estimate the conditional probability distribution $P(y|x)$ for each value of $x$ instead of producing a point estimate like $\mathbb{E}[y|x]$. This is called \emph{probabilistic regression}. Probabilistic regression is increasingly being used in fields like meteorology and healthcare \citep{gneiting_strictly_2007,avati_countdown_2019}. 

Probabilistic estimation is already the norm in \emph{classification} problems. Although some classifiers (e.g. standard support vector machines) only return a predicted class label, most are capable of returning estimated probabilities for each class; effectively, a conditional probability mass function. 

However, existing methods for probabilistic \emph{regression} are either inflexible, slow, or inaccessible to non-experts. Any mean-estimating regression method can be made probabilistic by assuming homoscedasticity and estimating an unconditional noise model, but homoscedasticity is a strong assumption and the process requires some statistical know-how. Generalized Additive Models for Shape, Scale, and Location (GAMLSS) allow heteroscedasticity but are restricted to a pre-specified model form \citep{gamlss}. Bayesian methods naturally generate predictive uncertainty estimates by integrating predictions over the posterior, but exact solutions to Bayesian models are limited to simple models, and calculating the posterior distribution of more powerful models such as Neural Networks (NN) \citep{Neal:1996:BLN:525544} and Bayesian Additive Regression Trees (BART) \citep{chipman_bart:_2010} is difficult. Inference in these models requires computationally expensive approximation via, for example, MCMC sampling. Moreover, sampling-based inference requires some statistical expertise and thus limits the ease-of-use of Bayesian methods. Bayesian approaches often also scale poorly to large datasets \citep{rasmussen_gaussian_2005}. Bayesian Deep Learning is gaining popularity \citep{graves_practical_2011, blundell_weight_2015, hernandez-lobato_probabilistic_2015} but, while neural networks have empirically excelled at perception tasks (such as with visual and audio input), they usually perform only on par with traditional methods when data are limited in size or tabular. Extensive hyper-parameter tuning and informative prior specification are also challenges for Bayesian Deep Learning which make it difficult to use out-of-the-box.

Meanwhile, Gradient Boosting Machines (GBMs) \citep{friedman_greedy_2001,chen_xgboost:_2016} are a set of highly-modular methods that work out-of-the-box and perform well on structured input data, even with relatively small datasets. This can be seen in their empirical success on Kaggle and other data science competitions \citep{chen_xgboost:_2016}. In classification tasks, their predictions are probabilistic by default (by use of the sigmoid or softmax link function). But in regression tasks, they output only a scalar value. Under a squared-error loss these scalars can be interpreted as the mean of a conditional Gaussian distribution with some (unknown) constant variance. However, such probabilistic interpretations have little use if the variance is assumed constant. The predicted distributions need to have at least two degrees of freedom (two parameters) to effectively convey both the magnitude and the uncertainty of the predictions, as illustrated in Figure \ref{fig:toy}. It is precisely this problem of simultaneous boosting of multiple parameters from the base learners which makes probabilistic forecasting with GBMs a challenge, and NGBoost addresses this with a multiparameter boosting approach and the use of natural gradients \citep{amari_natural_1998}.

\section{Summary of Contributions} \begin{enumerate}[i.]
    \item We present Natural Gradient Boosting, a modular algorithm for probabilistic regression (section \ref{ngb}) which uses multiparameter boosting and natural gradients to integrate any choice of:\vspace{-0.1cm} \begin{itemize}\itemsep0em \item Base learner (e.g. Regression Tree), \item Parametric probability distribution (Normal, Laplace, etc.), and \item Scoring rule (MLE, CRPS, etc.). \end{itemize}
    \item We present a generalization of the natural gradient to other scoring rules such as CRPS (section \ref{natgrad}).
    \item We demonstrate empirically that NGBoost performs competitively relative to other models in its predictive uncertainty estimates, as well as in its point estimates (section \ref{experiments}).
\end{enumerate}

\section{Natural Gradient Boosting}

In standard prediction settings, the object of interest is an estimate of a scalar function like $\mathbb{E}[y|x]$, where $x$ is a vector of observed features and $y$ is the prediction target. In our setting we are interested in producing a probability distribution $P_\theta(y|x)$ (with CDF $F_\theta$). Our approach is to assume $P_\theta(y|x)$ is of a specified parametric form, then estimate the $p$ parameters $\theta \in \mathbb{R}^p$ of the distribution as functions of $x$.

\subsection{Proper Scoring Rules}
To begin, we need a learning objective. In point prediction, the predictions are compared to the observed data with a loss function. The analogue in probabilistic regression is a \emph{scoring rule}, which compares the estimated probability distribution to the observed data.

A \emph{proper} scoring rule $\mathcal{S}$ takes as input a forecasted probability distribution $P$ and one observation $y$ (outcome), and assigns a score $\mathcal{S}(P, y)$ to the forecast such that the true distribution of the outcomes gets the best score in expectation \citep{gneiting_strictly_2007}. In mathematical notation, a scoring rule $\mathcal{S}$ is a proper scoring rule if and only if it satisfies
\begin{equation}
\E_{y\sim Q}[\mathcal{S}(Q, y)] \quad\le\quad \E_{y\sim Q}[\mathcal{S}(P, y)]\quad \forall P,Q \label{eqn:properineq},
\end{equation}
where $Q$ represents the true distribution of outcomes $y$, and $P$ is any other distribution (such as the probabilistic forecast from a model). Since we are working with parametric distributions, we can identify each distribution with its parameters and write the score as $\mathcal{S}(\theta, y)$.

The most commonly used proper scoring rule is the logarithmic score $\mathcal{L}$, which, when minimized, gives the MLE:
\begin{equation}
\mathcal{L}(\theta, y) = -\log P_\theta(y). 
\end{equation}
Another example is CRPS, which is generally considered a robust alternative to MLE \citep{gebetsberger_estimation_2018}. The CRPS (denoted $\mathcal{C}$) is defined as
\begin{equation}
\mathcal{C}(\theta, y) = \int_{-\infty}^y F_\theta(z)^2 dz + \int_y^\infty (1-F_\theta(z))^2 dz.
\end{equation}

\subsection{The Generalized Natural Gradient} \label{natgrad}

\begin{figure*}
    \centering
    
    \begin{subfigure}[t]{\linewidth}\begin{center}
        \includegraphics[width=0.7 \linewidth]{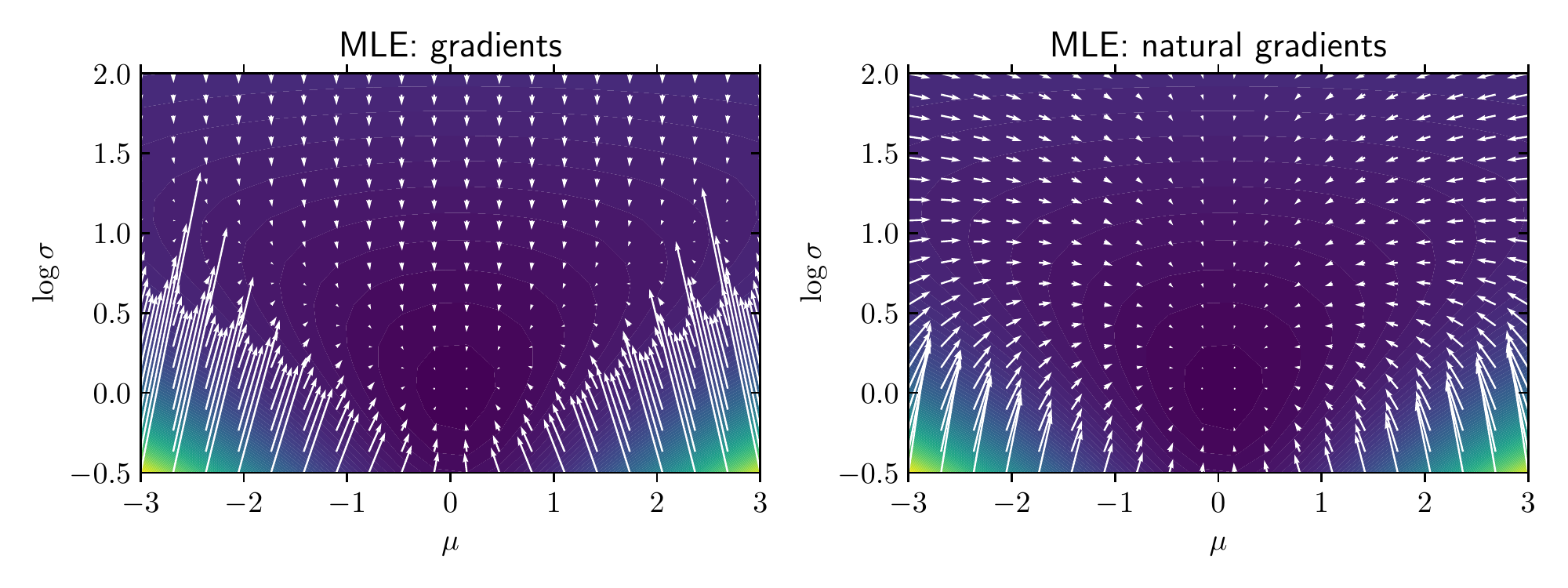} \end{center}
    \end{subfigure}
    \begin{subfigure}[t]{\linewidth}\begin{center}
        \includegraphics[width=0.7 \linewidth]{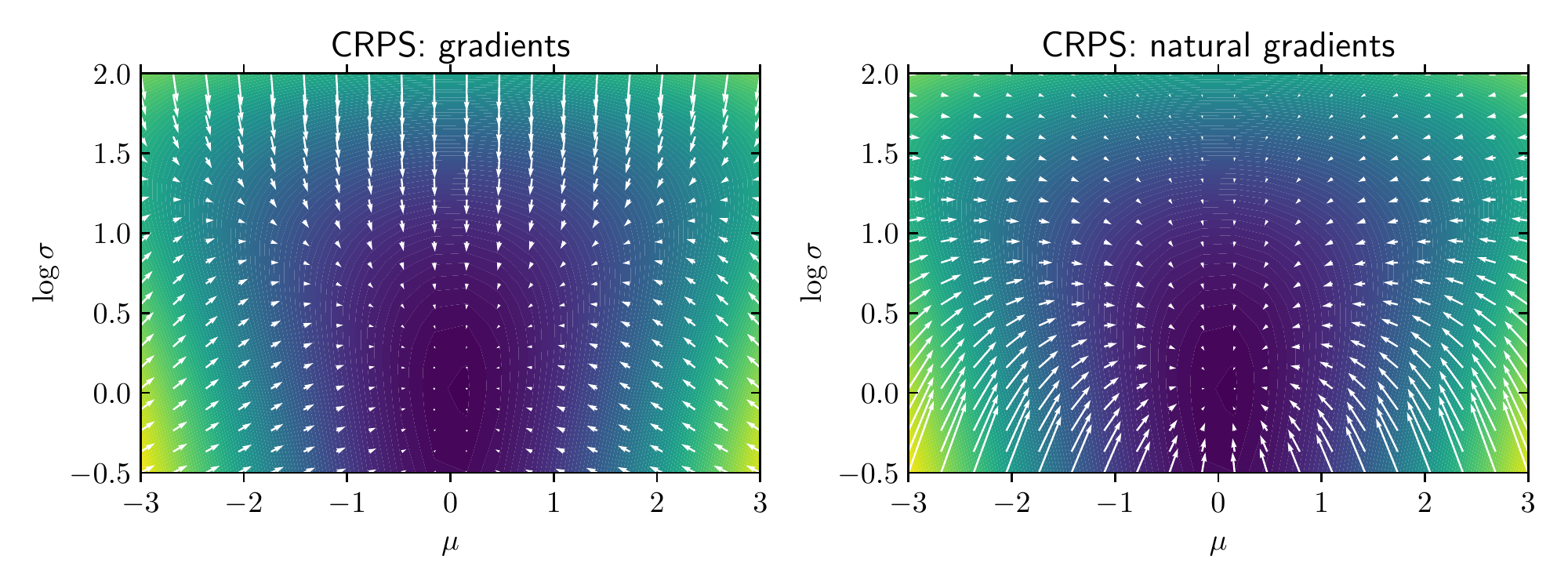}  \end{center}
    \end{subfigure}

    \caption{Proper scoring rules and corresponding gradients for fitting a Normal distribution on samples $\sim N(0,1)$. For each scoring rule, the landscape of the score (colors and contours) is identical, but the gradient fields (arrows) are markedly different depending on which kind of gradient is used.}
  
    \label{fig:gradients}
\end{figure*}

We take a standard gradient descent approach to find the parameters that minimize the scoring rule by descending along the negative gradient of the score relative to the parameters at each point $x$. The (ordinary) gradient of a scoring rule $\mathcal{S}$ over a parameterized probability distribution $P_\theta$ with parameter $\theta$ and outcome $y$ with respect to the parameters is denoted $\nabla \mathcal{S} (\theta, y)$.  It is the direction of steepest ascent, such that moving the parameters an infinitesimally small amount in that direction of the gradient (as opposed to any other direction) will increase the scoring rule the most. That is,
\begin{align}
\nabla \mathcal{S}(\theta, y) &\propto \lim_{\epsilon\to 0}\underset{d:\|d\|=\epsilon}{\arg\max} \mathcal{S}(\theta+d, y).
\end{align}
This gradient is \emph{not} invariant to reparameterization. Consider reparameterizing $P_\theta$ to $P_{z(\theta)}(y)$ so $P_\theta(y \in A) = P_\psi(y \in A)$ for all events $A$ when $\psi = z(\theta)$. If the gradient is calculated relative to $\theta$ and an infinitesimal step is taken in that direction, say from $\theta$ to $\theta + d\theta$ the resulting distribution will be different than if the gradient had been calculated relative to $\psi$ and a step was taken from $\psi$ to $\psi+d\psi$. In other words, $P_{\theta+d\theta}(y \in A) \ne P_{\psi+d\psi}(y \in A)$. Thus the choice of parameterization can drastically impact the training dynamics, even though the minima are unchanged.

The problem is that ``distance'' between two parameter values does not correspond to an appropriate ``distance'' between the distributions that those parameters identify. This motivates the natural gradient (denoted $\tilde{\nabla}$), which originated in information geometry  \citep{amari_natural_1998}. 

\paragraph{Divergences.} Every proper scoring rule induces a \emph{divergence} that can serve as local distance metric in the space of distributions. A proper scoring rule by definition satisfies the inequality of Eqn \ref{eqn:properineq}. The excess score of the right hand side over the left is the divergence induced by that scoring rule \citep{dawid_theory_2014}:
\begin{equation}
D_\mathcal{S}(Q \| P) = \mathbb{E}_{y\sim Q}[\mathcal{S}(P,y)] - \mathbb{E}_{y\sim Q}[\mathcal{S}(Q,y)],
\end{equation} 
which is necessarily non-negative, and can be interpreted as a measure of difference from one distribution $Q$ to another $P$. The MLE scoring rule induces the Kullback-Leibler divergence (KL divergence, or $D_{KL}$), while CRPS induces the $L^2$ divergence \citep{dawid_geometry_2007, machete_contrasting_2013}.

The divergences $D_{KL}$ and $D_{L^2}$ are invariant to how $Q$ and $P$ are parameterized. Though divergences in general are not symmetric, for small changes of the parameters they are almost symmetric and can serve as a local distance metric. When used as such, a divergence induces a statistical manifold where each point in the manifold corresponds to a probability distribution \citep{dawid_theory_2014}.

\paragraph{Natural Gradient.} The generalized natural gradient is the direction of steepest ascent in Riemannian space, which is invariant to parametrization, and is defined as:
\begin{equation}\label{natgradopt}
\tilde{\nabla}\mathcal{S}(\theta, y) \propto \lim_{\epsilon\to 0} \underset{d:D_{\mathcal S}(P_\theta||P_{\theta+d}) = \epsilon}{\arg\max} \mathcal{S}(\theta+d, y).
\end{equation}

If we solve the corresponding optimization problem, we obtain the natural gradient of the form
\begin{equation} \tilde{\nabla}\mathcal{S}(\theta, y) \quad\propto\quad \mathcal{I}_\mathcal{S}(\theta)^{-1}\nabla\mathcal{S}(\theta, y) \label{eqn:genericnatgrad}
\end{equation}
where $\mathcal{I}_\mathcal{S}(\theta)$ is the  Riemannian metric of the statistical manifold at $\theta$, which is induced by the scoring rule $\mathcal{S}$. While the natural gradient was originally defined for the statistical manifold with the distance measure induced by $D_{KL}$ \citep{martens_new_2014}, we provide a more general treatment here that applies to any divergence that corresponds to some proper scoring rule. 

By choosing $\mathcal{S} = \mathcal{L}$ (i.e. MLE) and solving the above optimization problem, we get:
\begin{equation}\label{natgradoptsol}
 \tilde\nabla \mathcal{L}(\theta, y) \quad\propto\quad \mathcal{I}_{\mathcal{L}}(\theta)^{-1} \nabla \mathcal{L}(\theta, y)
\end{equation}
where $\mathcal{I}_{\mathcal{L}}(\theta)$ is the Fisher Information carried by an observation about $P_\theta$, which is defined as:
\begin{align}
    \mathcal{I}_{\mathcal{L}}(\theta) &= \E_{ y\sim P_\theta}\left[\nabla_\theta \mathcal{L}(\theta, y) \nabla_\theta \mathcal{L}(\theta, y)^T \right] 
\end{align}

Similarly, by choosing $\mathcal{S} = \mathcal{C}$ (i.e. CRPS) and solving the above optimization problem, we get:
\begin{equation}\label{natgradcrpssol}
 \tilde\nabla \mathcal{C}(\theta, y) \quad\propto\quad \mathcal{I}_{\mathcal{C}}(\theta)^{-1} \nabla \mathcal{C}(\theta, y)
\end{equation}
where $\mathcal{I}_{\mathcal{C}}(\theta)$ is the Riemannian metric of the statistical manifold that uses $D_{L^2}$ as the local distance measure, given by \citep{dawid_geometry_2007}:
\begin{align}
    \mathcal{I}_{\mathcal{C}}(\theta) &= 2\int_{-\infty}^\infty \nabla_\theta F_\theta(z) \nabla_\theta F_\theta(z)^T dz.
\end{align}

Using the natural gradient for learning the parameters makes the optimization problem invariant to parametrization and leads to more efficient and stable learning dynamics  \citep{amari_natural_1998}. Figure \ref{fig:gradients} shows the vector field of gradients and natural gradients for $\mathcal{L}$ and $\mathcal{C}$ on the parameter space of a Normal distribution parameterized by $\mu$ (mean) and $\log\sigma$ (logarithm of the standard deviation).

\subsection{Gradient Boosting} \label{gbm}

Gradient boosting \citep{friedman_greedy_2001} is a supervised learning technique where several weak learners (or base learners) are combined in an additive ensemble. The model is learnt sequentially, where the next base learner is fit against the training objective residual of the current ensemble. The output of the fitted base learner is then scaled by a learning rate and added into the ensemble. 

The boosting framework can be generalized to any choice of base learner but most popular implementations use shallow decision trees because they work well in practice \citep{chen_xgboost:_2016,ke_lightgbm:_2017}.

When fitting a decision tree to the gradient, the algorithm partitions the data into axis-aligned slices. Each slice of the partition is associated with a leaf node of the tree, and is made as homogeneous in its response variable (the gradients at that set of data points) as possible. The criterion of homogeneity is typically the sample variance. The prediction value of the leaf node (which is common to all the examples ending up in the leaf node) is then set to be the additive component to the predictions that minimizes the loss the most. This is equivalent to doing a ``line search'' in the functional optimization problem for each leaf node, and, for some losses, closed form solutions are available. For example, for squared error, the response variables are residuals, and the result of the line search will yield the sample mean of the response variables in the leaf.

We now consider adapting gradient boosting for prediction of parameters $\theta$ in the probabilistic regression context.

\subsection{NGBoost: Natural Gradient Boosting} \label{ngb}

The NGBoost algorithm is a supervised learning method for probabilistic prediction that uses boosting to estimate the parameters of a conditional probability distribution $P(y|x)$ as functions of $x$. Here $y$ could be one of several types ($\{\pm1\}$, $\mathbb{R}$, $\{1,\ldots,K\}$, $\mathbb{R}_+$, $\mathbb{N}$, etc.) and $x$ is a vector in $\mathbb{R}^d$. In our experiments we focus on real valued outputs, though all of our methods are applicable to other modalities such as classification and survival prediction.

The algorithm has three modular components, which are chosen upfront as a configuration:
\vspace{-0.3cm}
\begin{itemize}\itemsep0em
    \item Base learner ($f$),
    \item Parametric probability distribution ($P_\theta$), and
    \item Proper scoring rule ($\mathcal{S}$).
\end{itemize}

\begin{algorithm}
\KwData{Dataset $\mathcal{D} = \{x_i,y_i\}_{i=1}^n$.}
\KwIn{Boosting iterations $M$, Learning rate $\eta$, Probability distribution with parameter $\theta$,  Proper scoring rule $\mathcal{S}$, Base learner $f$.}
 \KwOut{Scalings and base learners $\{\rho^{(m)},f^{(m)}\}_{m=1}^M.$ \vspace{0.02cm}}
$\theta^{(0)} \gets \arg\min_{\theta} \sum_{i=1}^{n}\mathcal{S}(\theta, y_i)$ \algorithmiccomment{initialize to marginal}\;

\For{$m \gets 1,\hdots,M$}{
    \For{$i \gets 1,\hdots,n$}{
         $g_i^{(m)} \gets \mathcal{I}_{\mathcal{S}}\left(\theta_i^{(m-1)}\right)^{-1}{\nabla_\theta} \mathcal{S}\left(\theta_i^{(m-1)}, {y}_i\right)$ \; 
    }
    
    $f^{(m)} \gets \mathsf{fit}\left(\left\{ {x}_i, g_i^{(m)} \right\}_{i=1}^{{n}}\right)$ \;
    
    $\rho^{(m)} \gets \arg\min_\rho \sum_{i=1}^{{n}}\mathcal{S}\left( \theta_i^{(m-1)} - \rho \cdot f^{(m)}({x}_i), {y}_i \right)$

     \For{$i \gets 1,\hdots,n$}{
        $\theta_i^{(m)} \gets \theta_i^{(m-1)} - \eta \left(  \rho^{(m)}\cdot f^{(m)}({x}_i)\right)$\;
    }
}
\caption{NGBoost for probabilistic prediction} 
\label{alg}
\end{algorithm}

A prediction $y|x$ on a new input $x$ is made in the form of a conditional distribution $P_\theta$, whose parameters $\theta$ are obtained by an additive combination of $M$ base learner outputs (corresponding to the $M$ gradient boosting stages) and an initial $\theta^{(0)}$. Note that $\theta$ can be a vector of parameters (not limited to be scalar valued), and they completely determine the probabilistic prediction $y|x$. For example, when using the Normal distribution, $\theta = (\mu, \log\sigma)$ in our experiments. To obtain the predicted parameter $\theta$ for some $x$, each of the base learners $f^{(m)}$ take $x$ as their input. Here $f^{(m)}$ collectively refers to the set of base learners, one per parameter, of stage $m$. For example, for a Normal distribution with parameters $\mu$ and $\log\sigma$, there will be two base learners, $f_\mu^{(m)}$ and $f_{\log\sigma}^{(m)}$ per stage, collectively denoted as $f^{(m)} = \left(f_\mu^{(m)}, f_{\log\sigma}^{(m)}\right)$. The predicted outputs are scaled with stage-specific scaling factors $\rho^{(m)}$, and a common learning rate $\eta$:
\[y|x \sim P_\theta(x), \quad\quad \theta = \theta^{(0)} - \eta\sum_{m=1}^M \rho^{(m)}\cdot f^{(m)}(x).\]
Each scaling factor $\rho^{(m)}$ is a single scalar, even if the distribution has multiple parameters. The model is learnt sequentially, a set of base learners $f^{(m)}$ and a scaling factor $\rho^{(m)}$ per stage.  The learning algorithm starts by first estimating a common $\theta^{(0)}$ such that it minimizes the sum of the scoring rule $\mathcal{S}$ over the response variables from all training examples, essentially fitting the marginal distribution of $y$. This becomes the initial predicted parameter $\theta^{(0)}$ for all examples.

In each iteration $m$, the algorithm calculates, for each example $i$, the  natural gradients $g_i^{(m)}$ of the scoring rule $\mathcal{S}$ with respect to the predicted parameters of that example up to that stage, $\theta^{(m-1)}_i$. Note that $g_i^{(m)}$ has the same dimension as $\theta$. A set of base learners for that iteration $f^{(m)}$ are fit to predict the corresponding components of the natural gradients $g_i^{(m)}$ of each example $x_i$.

The output of the fitted base learner is the projection of the natural gradient on to the range of the base learner class. This projected gradient is then scaled by a scaling factor $\rho^{(m)}$ since local approximations might not hold true very far away from the current parameter position. The scaling factor is chosen to minimize the overall true scoring rule loss along the direction of the projected gradient in the form of a line search. In practice, we found that implementing this line search by successive halving of $\rho$ (starting with $\rho=1$) until the scaled gradient update results in a lower overall loss relative to the previous iteration works reasonably well and is easy to implement.

Once the scaling factor $\rho^{(m)}$ is determined, the predicted per-example parameters are updated to $\theta^{(m)}_i$ by adding to each $\theta^{(m-1)}_i$ the negative scaled projected gradient for $i$, $\rho^{(m)}\cdot f^{(m)}(x_i)$ which is further scaled by a small learning rate $\eta$ (typically 0.1 or 0.01). 

The pseudo-code is presented in Algorithm \ref{alg}. For very large datasets computational performance can be easily improved by simply randomly sub-sampling mini-batches within the $\mathsf{fit}()$ operation.

\subsection{Analysis and Discussion} \label{sec:qualanal}

\begin{figure*}[t]
    \centering
    \begin{subfigure}[t]{0.24\linewidth}
        \begin{center}
        \caption{0\% fit}
        \includegraphics[width=\linewidth]{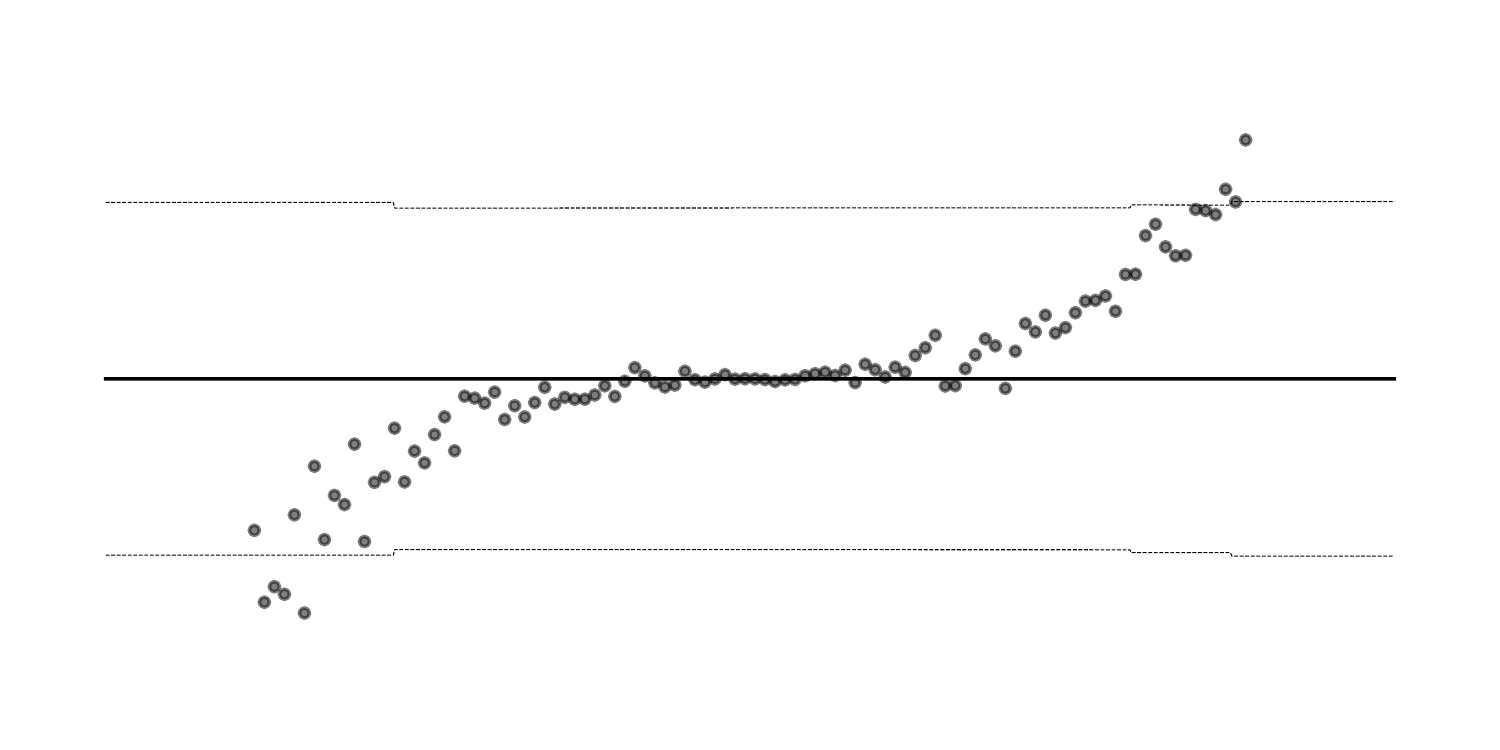} 
        \includegraphics[width=\linewidth]{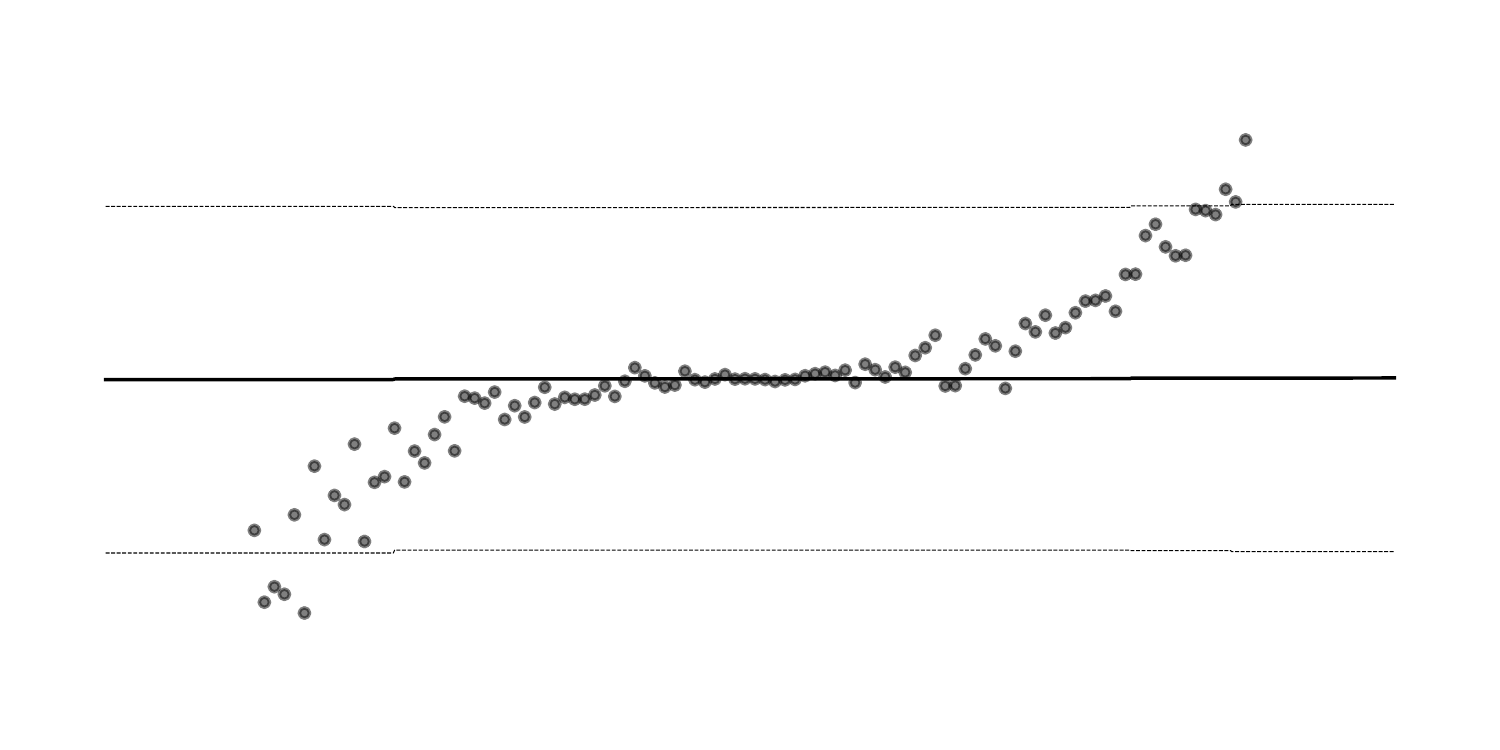} 
        \end{center}
    \end{subfigure}
    \begin{subfigure}[t]{0.24\linewidth}
        \begin{center}
        \caption{33\% fit}
        \includegraphics[width=\linewidth]{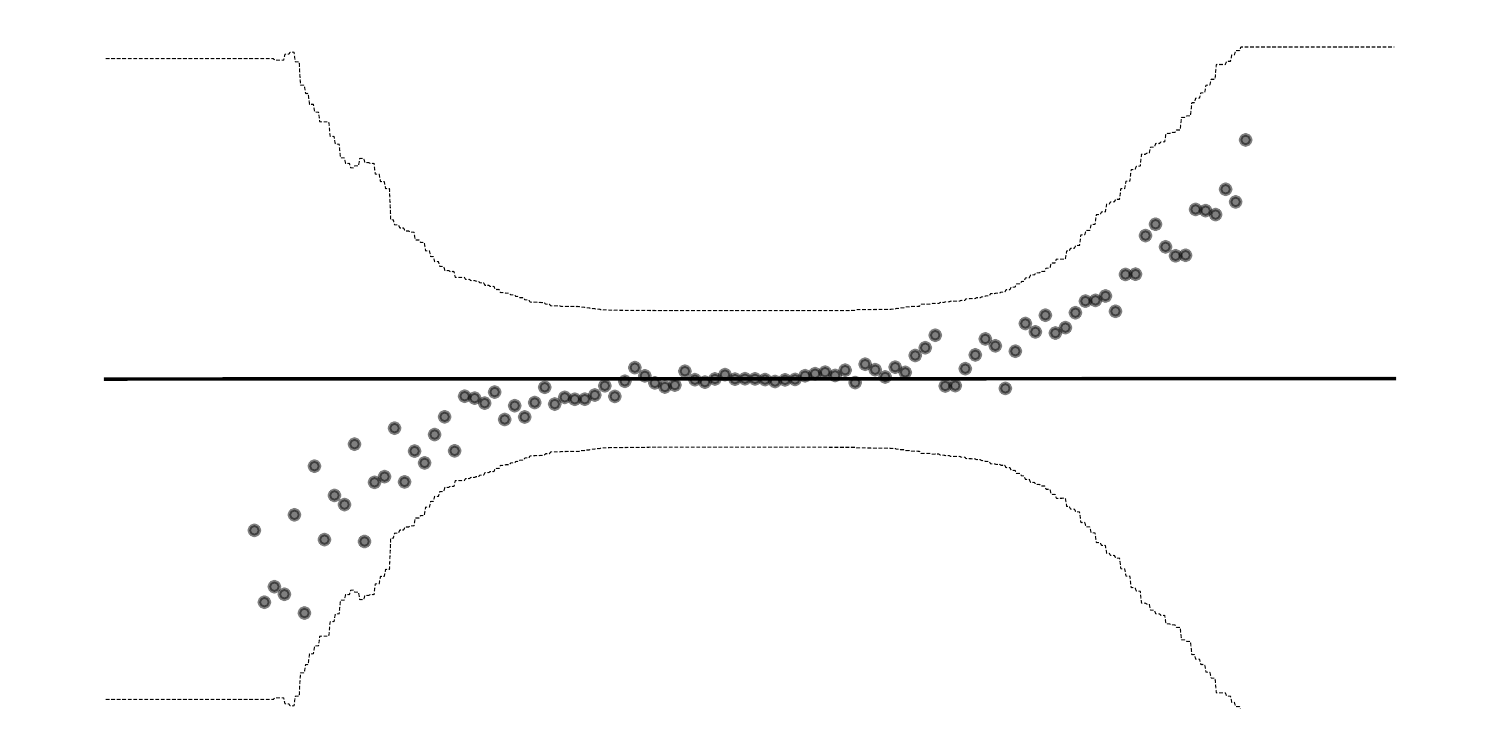} 
        \includegraphics[width=\linewidth]{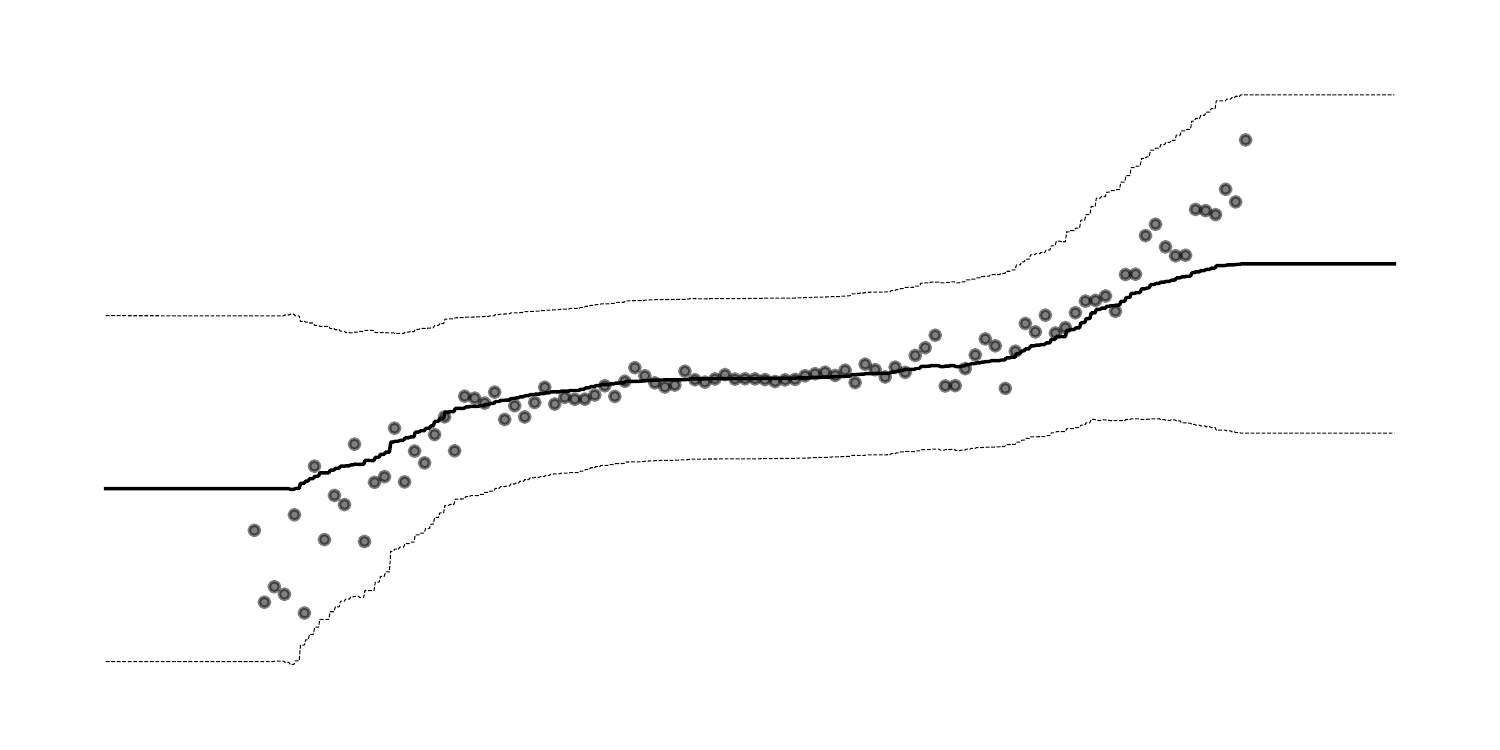} 
        \end{center}
    \end{subfigure}
    \begin{subfigure}[t]{0.24\linewidth}
        \begin{center}
        \caption{67\% fit}
        \includegraphics[width=\linewidth]{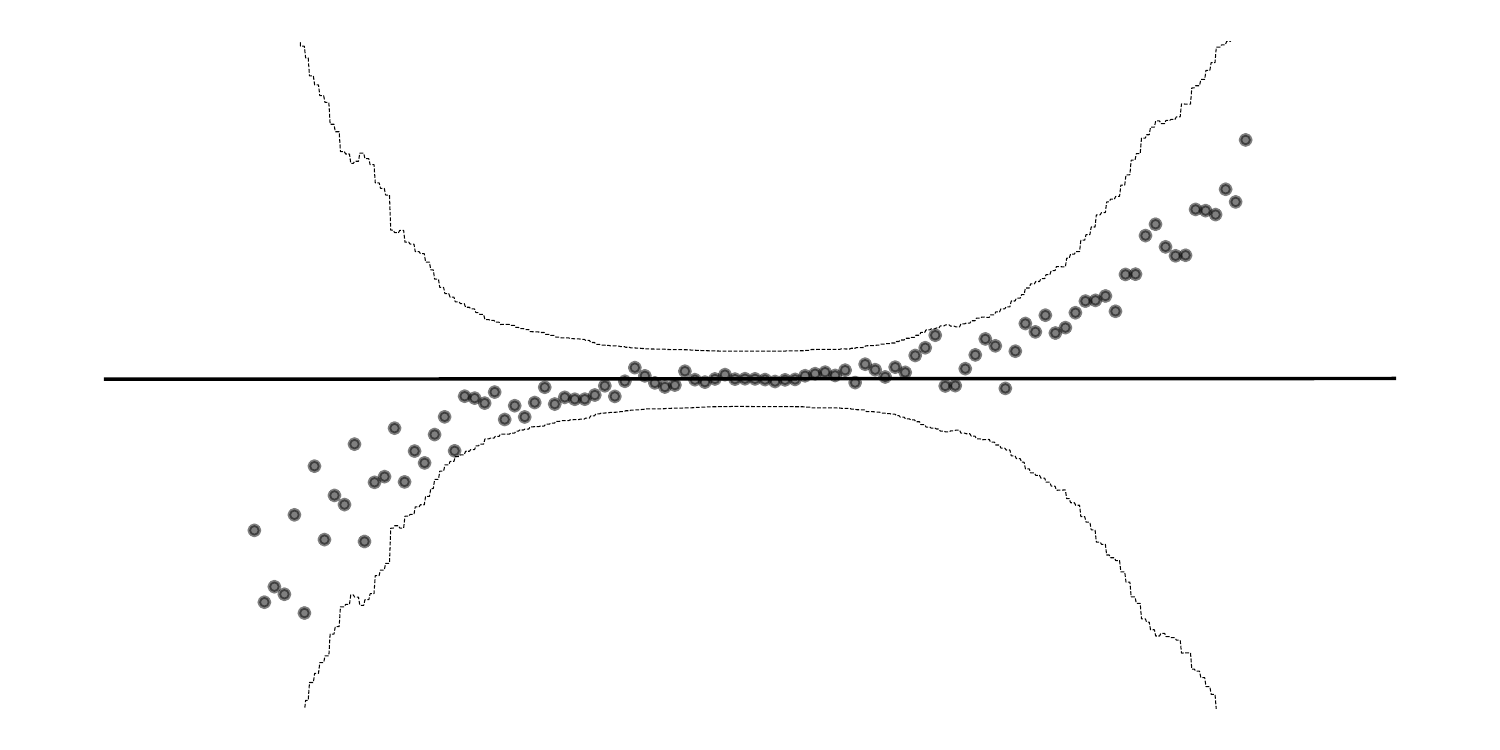} 
        \includegraphics[width=\linewidth]{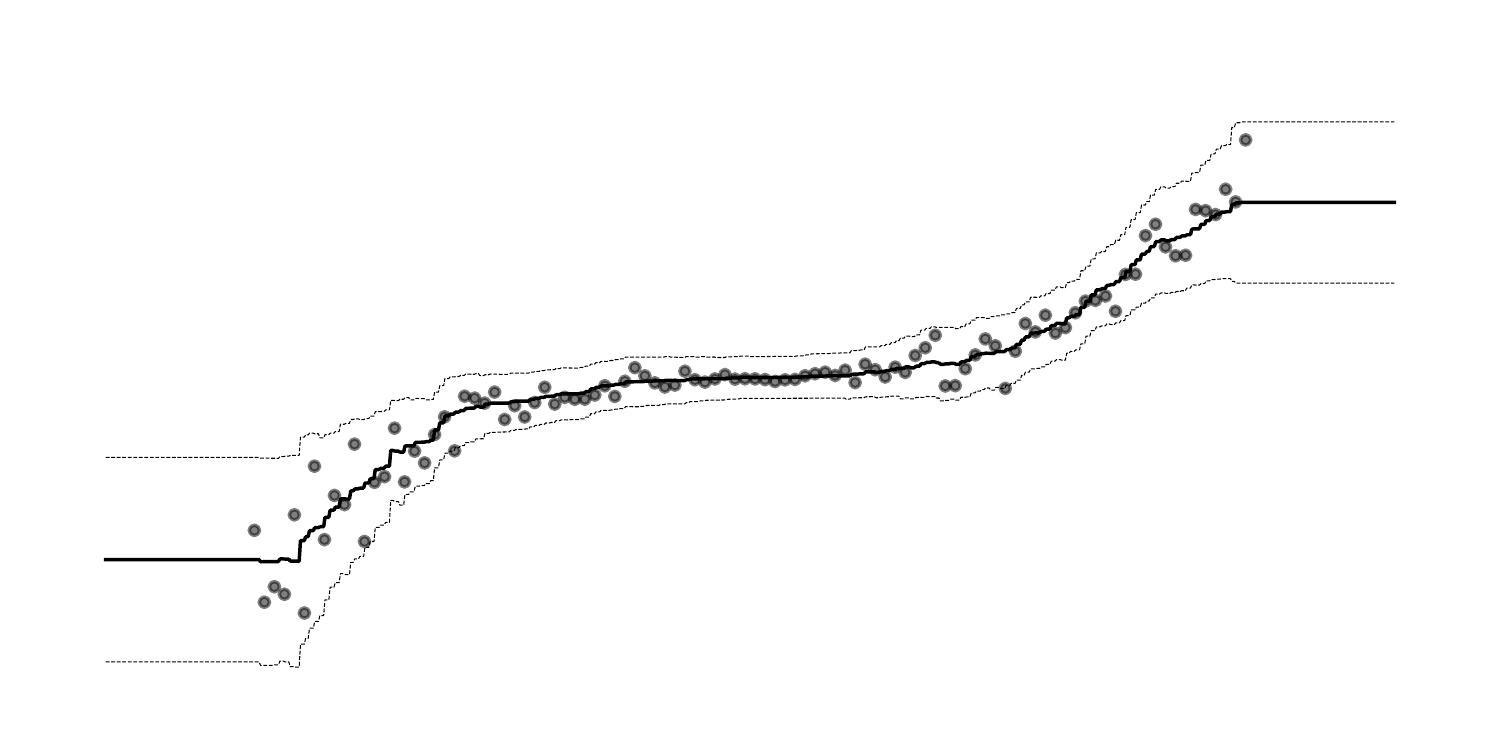} 
        \end{center}
    \end{subfigure}
    \begin{subfigure}[t]{0.24\linewidth}
        \begin{center}
        \caption{100\% fit}
        \includegraphics[width=\linewidth]{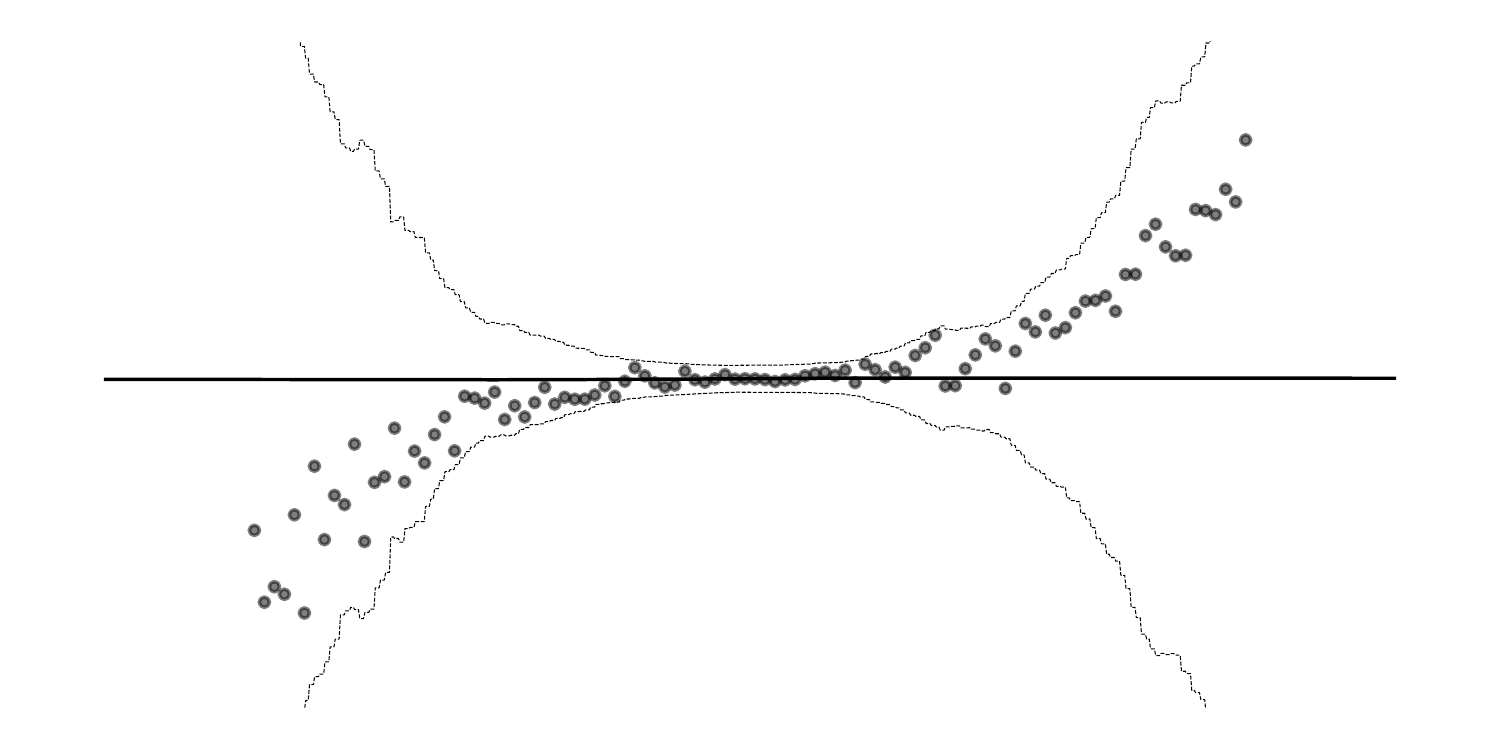} 
        \includegraphics[width=\linewidth]{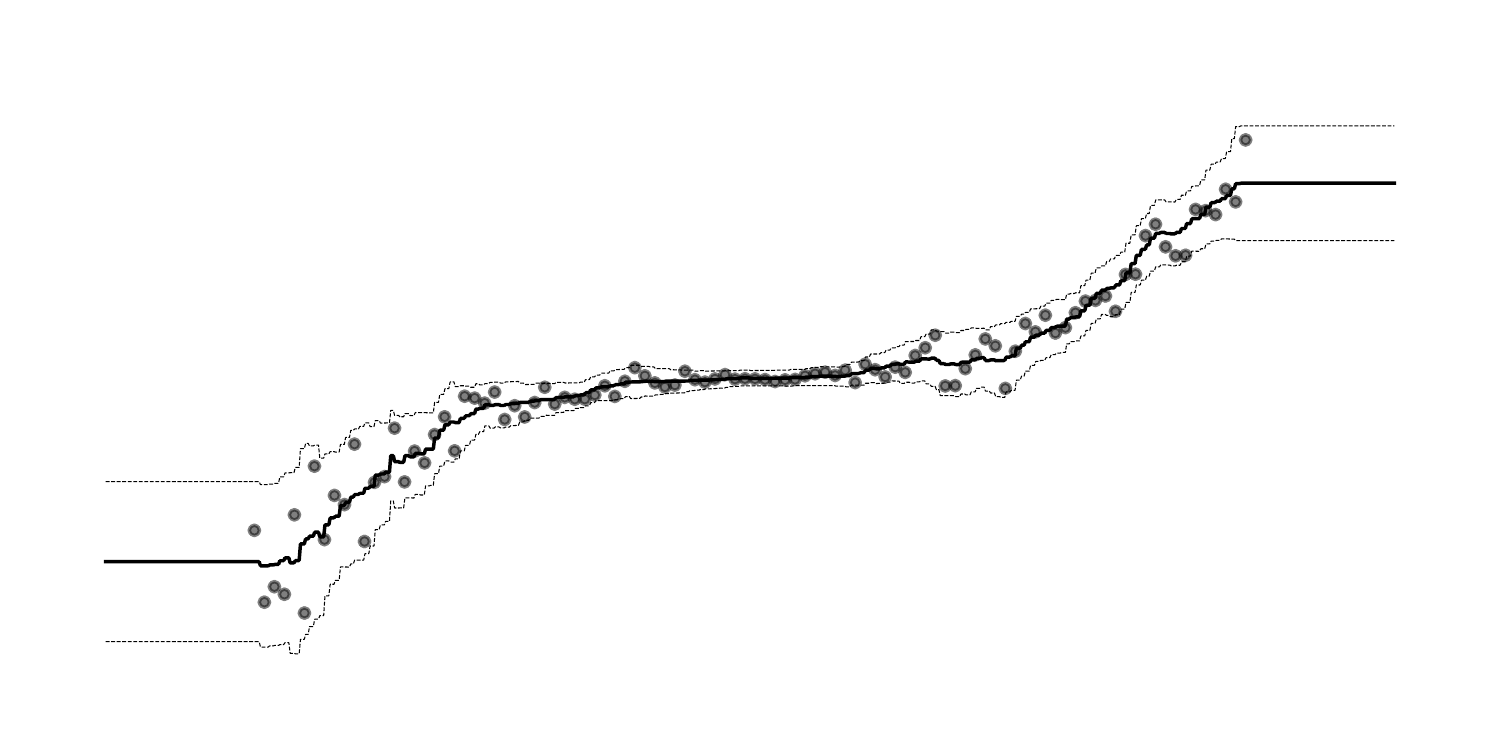} 
        \end{center}
    \end{subfigure}
    \caption{Contrasting the learning dynamics between using the ordinary gradient (top row) vs. the natural gradient (bottom row) for the purpose of gradient boosting the parameters of a Normal distribution on a toy data set. With ordinary gradients, we observe that ``lucky'' examples that are accidentally close to the initial predicted mean dominate the learning. This is because, under the ordinary gradient, the variances of those examples that have the correct mean get adjusted much more aggressively than the wrong means of the ``unlucky'' examples. This results in simultaneous overfitting of the ``lucky'' examples in the middle and underfitting of the ``unlucky'' examples at the ends. Under the natural gradient, all the updates are better balanced.}
    \label{fig:natvsnaive}
\end{figure*}

\paragraph{Boosting for probabilistic prediction.} 
Our boosting approach generalizes gradient boosting to predict conditional distributions. For instance, if the user specifies the conditional distribution to be a Normal distribution with a \emph{fixed} variance and uses the logarithmic scoring rule, our approach recovers the standard boosting algorithm with MSE loss (modulo the per-leaf line search). The advantage of NGBoost is that users are also free to specify any other family of distributions identified by a set of real-valued parameters and allow all of those parameters to vary over the covariates, not just the mean. NGBoost thus trivially extends to a variety of use cases, such as negative binomial boosting (for counts), Gamma or Weibull boosting (for survival prediction, with or without right-censored data), etc.

\paragraph{Multiparameter boosting.} These wide-ranging extensions of gradient boosting are made possible by turning the distributional prediction problem into a problem of jointly estimating $p$ functions of $x$, one per parameter, according to the scoring rule objective. In this setting, an overall line search for the stage multiplier (as opposed to per-leaf line search) is an inevitable consequence. However, our use of natural gradient makes this less of a problem as the gradients of all the examples come ``optimally pre-scaled'' (in both the relative magnitude between parameters, and across examples) due to the inverse Fisher Information factor. The use of ordinary gradients instead would be sub-optimal, as shown in Figure \ref{fig:natvsnaive}. With the natural gradient the parameters converge at approximately the same rate despite different conditional means, variances, and ``\emph{distances}'' from the initial marginal distribution, even while being subjected to a common scaling factor $\rho^{(m)}$ in each iteration. We attribute this stability to the ``optimal pre-scaling'' property of the natural gradient.

\paragraph{Parameterization.}
When the probability distribution is in the exponential family and the choice of parameterization is the natural parameters of that family, then a Newton-Raphson step is equivalent to a natural gradient descent step. However, in other parameterizations and distributions, the equivalence need not hold. This is especially important in the boosting context because, depending on the inductive biases of the base learners, certain parameterization choices may result in more suitable model spaces than others. For example, one setting we are particularly interested in is the two-parameter Normal distribution. Though it is in the exponential family, we use a mean ($\mu$) and log-scale ($\log\sigma$) parameterization for both ease of implementation and modeling convenience (to disentangle magnitude of predictions from uncertainty estimates). Since natural gradients are invariant to parameterization this does not pose a problem, whereas the Newton-Raphson method would fail as the problem is no longer convex in this parameterization. 

\paragraph{Computational complexity.}
There are two computational differences between our algorithm and a standard boosting algorithm which contribute to complexity. The first is that a series of learners must be fit for \textit{each} parameter in NGBoost, whereas standard boosting fits only one series of learners. The relative increase in computational cost is thus linear in the number of distributional parameters ($p$). The other difference is that we must compute the natural gradient per observation, which requires as many inversions of a $p \times p$ matrix $\mathcal I_{\mathcal S}$ as there are observations ($N$). The cost of doing so scales with $p^3$, and linearly with $N$. In practice, both costs are minimal because most commonly used distributions have only one or two parameters and distributions with more than five are exceedingly rare. Scaling in terms of $p$ is therefore not a significant concern. However, even though these costs are otherwise linear in $N$, it may be prudent to avoid inverting a large number of small matrices by sub-sampling mini-batches of data in each boosting iteration. This is done in most implementations of boosting algorithms. All in all, NGBoost scales exactly like other boosting algorithms in terms of $N$ but with larger ``constants'' that depend on $p \approx 10^0$. 

\section{Experiments} \label{experiments}

\begin{table*}[h]
\caption{Comparison of probabilistic regression performance on regression benchmark UCI datasets as measured by NLL. Results for MC dropout, Deep Ensembles, and Concrete Dropout are reported from \citet{gal_dropout_2016,lakshminarayanan_simple_2017,gal_concrete_2017} respectively. NGBoost offers competitive performance in terms of NLL, especially on smaller datasets. The best method for each dataset is bolded, as are those with standard errors that overlap with the best method.}
\vskip 0.1in
\small
\centering
\setlength\tabcolsep{3pt}
\begin{tabular}{lcccccccc}
\toprule
Dataset & $N$&  NGBoost                 & MC dropout              & Deep Ensembles              & Concrete Dropout      & Gaussian Process & GAMLSS & DistForest\\
\midrule
Boston & 506      &  \bf{2.43 $\pm$ 0.15}   &   \bf{2.46 $\pm$ 0.25}  &  \bf{2.41 $\pm$ 0.25}       & 2.72 $\pm$ 0.01       & \bf{2.37 $\pm$ 0.24} & 2.73 $\pm$ 0.56 & 2.67 $\pm$ 0.08\\
Concrete & 1030   &   \bf{3.04 $\pm$ 0.17}  &  \bf{3.04 $\pm$ 0.09}   &  \bf{3.06 $\pm$ 0.18}       & 3.51 $\pm$ 0.00       & \bf{3.03 $\pm$ 0.11} & 3.24 $\pm$ 0.08 & 3.38 $\pm$ 0.05\\
Energy & 768      & \bf{0.60  $\pm$ 0.45}   &  1.99 $\pm$ 0.09        &  1.38 $\pm$ 0.22            & 2.30 $\pm$ 0.00       & \bf{0.66 $\pm$ 0.17} & 1.24 $\pm$ 0.86 & 1.53 $\pm$ 0.14\\
Kin8nm & 8192     &   -0.49 $\pm$ 0.02      &  -0.95 $\pm$ 0.03       &  \bf{-1.20 $\pm$ 0.02}      & -0.65 $\pm$ 0.00      & -1.11 $\pm$ 0.03 & -0.26 $\pm$ 0.02 & -0.40 $\pm$ 0.01\\
Naval   & 11934   &  -5.34$\pm$ 0.04       & -3.80 $\pm$ 0.05        &  -5.63 $\pm$ 0.05           & \bf{-5.87 $\pm$ 0.05} & -4.98 $\pm$ 0.02 & -5.56 $\pm$ 0.07 & -4.84 $\pm$ 0.01\\
Power & 9568      & 2.79  $\pm$  0.11  &  2.80 $\pm$ 0.05   &  2.79 $\pm$ 0.04       & 2.75 $\pm$ 0.01  & 2.81 $\pm$ 0.05 & 2.86 $\pm$ 0.04 & \bf{2.68 $\pm$ 0.05}\\
Protein &  45730  &  2.81 $\pm$  0.03  &  2.89 $\pm$ 0.01        &  2.83 $\pm$ 0.02       & 2.81 $\pm$ 0.00  & 2.89 $\pm$ 0.02 & 3.00 $\pm$ 0.01 & \bf{2.59 $\pm$ 0.04}\\
Wine & 1588       &   \bf{0.91 $\pm$ 0.06}  &  \bf{0.93 $\pm$ 0.06}   &  \bf{0.94 $\pm$ 0.12}       & 1.70 $\pm$ 0.00       & \bf{0.95 $\pm$ 0.06} & \bf{0.97 $\pm$ 0.09} & 1.05 $\pm$ 0.15\\
Yacht & 308       & \bf{0.20  $\pm$ 0.26}   &  1.55 $\pm$ 0.12        &  1.18 $\pm$ 0.21            & 1.75 $\pm$ 0.00       & \bf{0.10 $\pm$ 0.26} & 0.80 $\pm$ 0.56 & 2.94 $\pm$ 0.09\\ 
Year MSD & 515345 &  3.43 $\pm$ NA          &  3.59 $\pm$ NA          &  \bf{3.35 $\pm$ NA}         & NA$\pm$ NA            & NA $\pm$ NA  & NA $\pm$ NA & NA $\pm$ NA\\
\bottomrule
\end{tabular}
\label{tab:NLLresults}
\end{table*}

\begin{table*}[h]
\caption{Comparison of probabilistic regression performance on regression benchmark UCI datasets as measured by NLL while ablating key components of NGBoost. Multiparameter boosting must be used in tandem with the natural gradient to increase performance. Bolding is as in Table 1.}
\vskip 0.1in
\small
\centering
\begin{tabular}{lccccc}
\toprule
Dataset       &     N     & NGBoost        & 2nd-Order         & Multiparameter              		  & Homoscedastic			              	 \\
\midrule
Boston & 506      & \bf{2.43 $\pm$ 0.15} 	& 3.57 $\pm$ 0.20				& 3.17 $\pm$ 0.13 &  \bf{2.79 $\pm$ 0.42}   \\
Concrete & 1030   & \bf{3.04 $\pm$ 0.17}  	& 4.21 $\pm$ 0.05				& 3.94 $\pm$ 0.09  & \bf{3.22 $\pm$ 0.29}        \\
Energy & 768      & \bf{0.60 $\pm$ 0.45}  	& 3.64 $\pm$ 0.06				& 3.24 $\pm$ 0.09 &    \bf{0.68 $\pm$ 0.25} \\
Kin8nm & 8192     & -0.49 $\pm$ 0.02  		& 0.10 $\pm$ 0.07				& \bf{-0.52 $\pm$ 0.03} &  -0.37 $\pm$ 0.05      \\
Naval   & 11934   & \bf{-5.34$\pm$ 0.04}  	& -2.80 $\pm$ 0.01				& -3.46 $\pm$ 0.00  & -4.35 $\pm$ 0.07 \\
Power & 9568      & 2.79 $\pm$ 0.11 		& 4.11 $\pm$ 0.03				& 3.79 $\pm$ 0.13 & \bf{2.66 $\pm$ 0.11}    \\
Protein &  45730  & \bf{2.81 $\pm$ 0.03}  	& 3.23 $\pm$ 0.00				& 3.04 $\pm$ 0.02  &   2.86 $\pm$ 0.01  \\
Wine & 1588       & \bf{0.91 $\pm$ 0.06}  	& 1.21 $\pm$ 0.09				& 0.93 $\pm$ 0.07  &   \bf{1.34 $\pm$ 0.67}  \\
Yacht & 308       & \bf{0.20 $\pm$ 0.26}  	& 4.11 $\pm$ 0.17	& 3.29 $\pm$ 0.20   &   2.02 $\pm$ 0.21  \\
Year MSD & 515345 & \bf{3.43 $\pm$ NA}  	& 3.80 $\pm$ 0.00				& 3.60 $\pm$ NA  & 3.63 $\pm$ NA\\
\bottomrule
\end{tabular}
\label{tab:ablationresults}
\end{table*}

\begin{table*}[h]
\caption{Comparison of point-estimation performance on regression benchmark UCI datasets as measured by RMSE. Although not optimized for point estimation, NGBoost still offers competitive performance. Bolding is as in Table \ref{tab:NLLresults}.}
\vskip 0.1in
\small
\centering
\begin{tabular}{lccccccc}
\toprule
  Dataset & $N$& NGBoost & Elastic Net & Random Forest & Gradient Boosting & GAMLSS & Distributional Forest\\
\midrule
  Boston    & 506    &  \bf{2.94 $\pm$ 0.53}  &   4.08 $\pm$ 0.16 & 2.97 $\pm$ 0.30 & \bf{2.46 $\pm$ 0.32} & 4.32 $\pm$ 1.40 & 3.99 $\pm$ 1.13\\
  Concrete  & 1030   &  \bf{5.06 $\pm$ 0.61}  &  12.1 $\pm$ 0.05 & 5.29 $\pm$ 0.16 & \bf{4.46 $\pm$ 0.29} & 6.72 $\pm$ 0.59 & 6.61 $\pm$ 0.83 \\
  Energy    & 768    & 0.46 $\pm$ 0.06   &  2.75 $\pm$ 0.03 & 0.52 $\pm$ 0.09 & \bf{0.39 $\pm$ 0.02} & 1.43 $\pm$ 0.32 & 1.11 $\pm$ 0.27\\
  Kin8nm    & 8192   & 0.16  $\pm$ 0.00       &  0.20 $\pm$ 0.00 & 0.15 $\pm$ 0.00 & \bf{0.14 $\pm$ 0.00} & 0.20 $\pm$ 0.01 & 0.16 $\pm$ 0.00\\
  Naval     & 11934  &  \bf{0.00 $\pm$ 0.00}  & \bf{0.00 $\pm$ 0.00} & \bf{0.00 $\pm$ 0.00} & \bf{0.00 $\pm$ 0.00} & \bf{0.00 $\pm$ 0.00} & \bf{0.00 $\pm$ 0.00}\\
  Power     & 9568   & 3.79  $\pm$ 0.18  &  4.42 $\pm$ 0.00 & 3.26 $\pm$ 0.03 & \bf{3.01 $\pm$ 0.10} & 4.25 $\pm$ 0.19 & 3.64 $\pm$ 0.24 \\
  Protein   &  45730 &  4.33 $\pm$ 0.03       & 5.20 $\pm$ 0.00 & \bf{3.60 $\pm$ 0.00} & 3.95 $\pm$ 0.00 & 5.04 $\pm$ 0.04 & 3.89 $\pm$ 0.04\\
  Wine      & 1588   & 0.63  $\pm$ 0.04  & 0.58 $\pm$ 0.00 & \bf{0.50 $\pm$ 0.01} & 0.53 $\pm$ 0.02 & 0.64 $\pm$ 0.04 & 0.67 $\pm$ 0.05\\
  Yacht     & 308    &  \bf{0.50  $\pm$ 0.20} & 7.65 $\pm$ 0.21 & 0.61 $\pm$ 0.08 & \bf{0.42 $\pm$ 0.09} & 8.29 $\pm$ 2.56 & 4.19$\pm$ 0.92\\ 
  Year MSD   & 515345 &   8.94 $\pm$ NA        & 9.49 $\pm$ NA & 9.05 $\pm$ NA & \bf{8.73 $\pm$ NA} & NA $\pm$ NA & NA $\pm$ NA\\
\bottomrule
\end{tabular}
\label{tab:RMSEresults}
\end{table*}

Our experiments use datasets from the UCI Machine Learning Repository, and follow the protocol first proposed in \citet{hernandez-lobato_probabilistic_2015}. For all datasets, we hold out a random 10\% of the examples as a test set. From the other 90\% we initially hold out 20\% as a validation set to select  $M$ (the number of boosting stages) that gives the best log-likelihood, and then retrain on the entire 90\% using the chosen $M$. The retrained model is then made to predict on the held-out 10\% test set. This entire process is repeated 20 times for all datasets except Protein and Year MSD, for which it is repeated 5 times and 1 time respectively.

For all experiments, NGBoost was configured with the Normal distribution, decision tree base learner with a maximum depth of three levels, and log scoring rule.  The Year MSD dataset, being extremely large relative to the rest, was fit using a learning rate $\eta$ of 0.1 while the rest of the datasets were fit with a learning rate of 0.01. In general we recommend small learning rates, subject to computational feasibility. For the Year MSD dataset we use a mini-batch size of 10\%, for all other datasets we use 100\%.

\subsection{Probabilistic regression.} The quality of predictive uncertainty is captured in the average negative log-likelihood (NLL) (i.e. $\log \hat{P_\theta}(y|x)$) as measured on the test set. 

Our comparison in this task is against other probabilistic prediction methods. Namely:

\textbf{MC dropout} fits a neural network to the dataset and interprets Bernoulli dropout as a variational approximation for Bayesian inference, obtaining predictive uncertainty by integrating over Monte Carlo samples \citep{gal_dropout_2016}. We use the results from \citet{gal_dropout_2016} as our benchmark.

\textbf{Deep Ensembles} fit an ensemble of neural networks to the dataset and obtain predictive uncertainty by making an approximation to the Gaussian mixture arising out of the ensemble \citep{lakshminarayanan_simple_2017}. We use the results from \citet{lakshminarayanan_simple_2017} as our benchmark.

\textbf{Concrete Dropout} improves upon MC dropout by employing a continuous relaxation of the Bernoulli distribution to automatically tune the dropout probability \citep{gal_concrete_2017}. We use the results from \citet{gal_concrete_2017} as our benchmark.

\textbf{Gaussian Processes} are a nonparametric Bayesian method where the response is interpreted as a multivariate Gaussian distribution with covariance given by some kernel between covariates \citep{rasmussen_gaussian_2005}. Our experiments used an automatic relevance detection kernel fit via gradient-based optimization of the marginal log-likelihood. Datasets with $N>2000$ employed 1000 inducing inputs randomly chosen from the training set, with inducing points fit with variational inference as in \citet{titsias_variational_2009}. All features and labels were standardized to zero-mean and unit variance for pre-processing. The standardized noise level was tuned via grid search for each dataset, with values ranging between 0.01 and 0.1. 

\textbf{GAMLSS} uses generalized (parametric) linear models to fit each distributional parameter instead of boosting  \citep{gamlss}. Responses were parameterized as Normal distributions $N(\mu,\sigma^2)$.  The mean $\mu$ and log-std $\log \sigma$ were independently modeled as linear combinations of natural cubic splines of the covariates. No interaction terms were included. All features and labels were standardized to zero-mean and unit variance for pre-processing. 

\textbf{Distributional Forests} use trees to estimate distributional parameters in each leaf, which are then averaged across the model \citep{schlosser_distributional_2019}. Responses were parameterized as Normal distributions $N(\mu,\sigma^2)$. The mean $\mu$ and log-std $\log \sigma$ were independently modeled using forests consisting of 200 trees and default hyper-parameters ($\sqrt d$ covariates sampled per split, minimum 20 examples for a split node, minimum 7 examples in a terminal node). All features and labels were standardized to zero-mean and unit variance for pre-processing. 

Our probabilistic regression results are summarized in Table \ref{tab:NLLresults}. Results for the Year MSD dataset are unavailable either because they were not reported or because the necessary computations for gradient-based optimization of hyper-parameters did not fit in memory. 

\subsection{Ablation} 

We compare the NLL of our full NGBoost algorithm on these data versus that of the following comparators, each tuned in the same fashion: 

\textbf{2nd-Order} boosting is NGBoost using 2nd-order gradient descent instead of the natural gradient. This tests the added benefit of using the natural gradient vis-a-vis 2nd-order methods. Recent work has argued that the natural gradient improves training dynamics by approximating the Hessian used in 2nd-order methods \cite{martens_new_2014}. We use the ``saddle-free'' Newton-Raphson method of \citet{dauphin_identifying_2014} in our implementation of 2nd-order multiparameter boosting to provide a strong baseline.

\textbf{Multiparameter} boosting is NGBoost using the ordinary gradient. This tests the added benefit of using the natural gradient vis-a-vis the standard gradient, but still allows for all of the parameters of the distribution to vary across $x$.

\textbf{Homoscedastic} boosting is NGBoost assuming a homoscedastic variance $\sigma^2(x) = \sigma^2 = \widehat{\text{Var}[r]}$ where $r$ are the training set residuals from a single-parameter (mean) boosting model. This tests the added benefit of allowing parameters other than the conditional mean to vary across $x$. Note that the natural gradient plays no meaningful role when there is only a single parameter estimated with NGBoost. 
    
Our ablation results are summarized in Table \ref{tab:ablationresults}. 

\subsection{Point estimation} 
Although NGBoost is not specifically designed for point estimation, it is easy to extract point estimates of expectations $\hat{\mathbb{E}}[y|x]$ from the estimated distributions $\hat P_\theta(y|x)$. We use this approach in a third evaluation to compare the same NGBoost models as above to the Scikit-Learn implementations of random forests, standard gradient boosting, and elastic net regression \citep{pedregosa_scikit-learn:_2011}. Predictive performance in this evaluation is captured by the root mean squared-error (RMSE) of the predictions on the test set. We performed hyperparameter tuning for each of the comparator methods using the same validation procedure as described above, although optimizing for RMSE instead of NLL for all methods except NGBoost. We compare to:

\textbf{Elastic Net}: We used the Scikit-Learn implementation of elastic net regularized linear models. We tuned over lasso-ridge mixture parameters of 0.01, 0.7, and 0.99 and over a range of regularization parameters between 0.00005 and 0.01. All other parameters were left to their default values.

\textbf{Random Forest}: We used the Scikit-Learn implementation of random forests. We set the number of trees to 500 and left all other parameters at their default values.

\textbf{Gradient Boosting}: We used the Scikit-Learn implementation of gradient boosted trees. We tuned over learning rates of 0.01, 0.05, and 0.1, tree depths of 3 and 4, and number of boosting iterations between 0 and 1000. All other parameters were left to their default values.

Our point prediction results are summarized in Table \ref{tab:RMSEresults}.

\section{Conclusions}

NGBoost is a method for probabilistic prediction with competitive state-of-the-art performance on a variety of datasets. NGBoost combines a multiparameter boosting algorithm with the natural gradient to efficiently estimate how parameters of the presumed outcome distribution vary with the observed features. 

NGBoost performs as well as existing methods for probabilistic regression but retains major advantages: NGBoost is flexible, scalable, and easy-to-use. We have not rigorously quantified these advantages in this paper (since they would be largely irrelevant without first establishing performance), but many of the benefits are self-evident. Unlike problem-specific approaches, NGBoost handles classification, regression, survival problems, etc. using the same software package and interface. NGBoost scales to large numbers of features or observations with the same favorable complexity of traditional boosting algorithms. No expert knowledge of deep learning, Bayesian statistics, or Monte Carlo methods is required to use NGBoost. It works out of the box.

Our ablation experiments demonstrate that multiparameter boosting and the natural gradient work together to improve performance. Assuming a uniform variance across all covariates works reasonably well for some datasets, as would be expected, but this is not always the case. However, using multiparameter boosting to relax the homoscedasticity assumption most often results in worse performance, likely due to poor training dynamics. 2nd-order methods result in even worse performance. NGBoost employs the natural gradient to correct the training dynamics of multiparameter boosting. The superiority to 2nd-order methods demonstrates that this is due to exploiting the curvature of the score in distributional space, not the curvature of the score in parameter space. The result is performance that is almost always better than assuming homoscedasticity, sometimes by a large margin.

Furthermore, the advantages of probabilistic regression come almost ``for free''. On point estimation tasks NGBoost performs better than elastic net, about on par with random forests, and within striking distance of gradient boosting. This is despite the fact that the NGBoost models were (a) optimized for NLL, not to minimize RMSE and (b) less aggressively tuned. Thus, although point prediction will always be best with a dedicated model for that purpose, the loss in RMSE is not substantial if NGBoost is used in order to support probabilistic regression instead.

There are many avenues for future work. This paper is focused on regression problems for clarity of exposition, but NGBoost is also applicable to classification and to survival problems with right-censored data (using the censored likelihood as a scoring rule). NGBoost could also be used for \emph{joint} prediction: by modeling two outcomes $z$ and $y$ with a jointly parameterized conditional distribution $P_{\theta}(z,y|x)$, a single NGBoost model could answer any question like ``what is the probability that it rains more than 4 inches \emph{and} the temperature is greater than 17\degree C tomorrow?''.

Some further technical innovations are also worth exploring. The natural gradient loses its invariance property with finite step sizes, which we can address with differential equation solvers for higher-order invariance \citep{song_accelerating_2018}. Better tree-based base learners and regularization (e.g. \citet{chen_xgboost:_2016, ke_lightgbm:_2017}) are also likely to improve performance, especially in terms of scaling to large datasets.

Although we have shown empirically that NGBoost is useful for probabilistic prediction, it remains to be seen whether it is useful for inference problems and under what assumptions. For instance, if we assume that $y|x \sim \mathcal D(\theta(x))$ for some distribution $\mathcal D$ with parameters $\theta$ and we estimate $\hat\theta_{\text{ngb}}(x)$ using NGBoost, under what conditions do we have that $\hat\theta_{\text{ngb}}(x) \rightarrow \theta(x)$ as sample size increases? Are there conditions where the convergence is uniform in $x$? If the model is misspecified (i.e. $\mathcal D$ is not correct), are there conditions under which moment estimates from the model are still consistent? Addressing these questions and others like them would be of significant value.

\subsection*{Acknowledgements}

This work was funded in part by the National Institutes of Health. We thank anonymous reviewers for feedback.

\bibliographystyle{apalike}
\bibliography{references.bib,url.bib}

\end{document}